\documentclass[10pt,a4paper,onecolumn]{article}
\usepackage{marginnote}
\usepackage{graphicx}
\usepackage{xcolor}
\usepackage{authblk,etoolbox}
\usepackage{titlesec}
\usepackage{calc}
\usepackage{tikz}
\usepackage{hyperref}
\hypersetup{colorlinks,breaklinks=true,
            urlcolor=[rgb]{0.0, 0.5, 1.0},
            linkcolor=[rgb]{0.0, 0.5, 1.0}}
\usepackage{caption}
\usepackage{tcolorbox}
\usepackage{amssymb,amsmath}
\usepackage{ifxetex,ifluatex}
\usepackage{seqsplit}
\usepackage{xstring}

\usepackage{float}
\let\origfigure\figure
\let\endorigfigure\endfigure
\renewenvironment{figure}[1][2] {
    \expandafter\origfigure\expandafter[H]
} {
    \endorigfigure
}

\usepackage{fixltx2e} % provides \textsubscript
\usepackage[
  backend=biber,
]{biblatex}
\bibliography{paper.bib}

\let\textttOrig=\texttt
\def\texttt#1{\expandafter\textttOrig{\seqsplit{#1}}}
\renewcommand{\seqinsert}{\ifmmode
  \allowbreak
  \else\penalty6000\hspace{0pt plus 0.02em}\fi}

\makeatletter
\let\href@Orig=\href
\def\href@Urllike#1#2{\href@Orig{#1}{\begingroup
    \def\Url@String{#2}\Url@FormatString
    \endgroup}}
\def\href@Notdoi#1#2{\def\tempa{#1}\def\tempb{#2}%
  \ifx\tempa\tempb\relax\href@Urllike{#1}{#2}\else
  \href@Orig{#1}{#2}\fi}
\def\href#1#2{%
  \IfBeginWith{#1}{https://doi.org}%
  {\href@Urllike{#1}{#2}}{\href@Notdoi{#1}{#2}}}
\makeatother

\newlength{\cslhangindent}
\newlength{\csllabelwidth}
\newenvironment{CSLReferences}[3] % #1 hanging-ident, #2 entry spacing
 {% don't indent paragraphs
  \setlength{\parindent}{0pt}
  \ifodd #1 \everypar{\setlength{\hangindent}{\cslhangindent}}\ignorespaces\fi
  \ifnum #2 > 0
  \setlength{\parskip}{#2\baselineskip}
  \fi
 }%
 {}
\usepackage{calc}

\usepackage[top=3.5cm, bottom=3cm, right=1.5cm, left=1.0cm,
            headheight=2.2cm, reversemp, includemp, marginparwidth=4.5cm]{geometry}

\titleformat{\section}
  {\normalfont\sffamily\Large\bfseries}
  {}{0pt}{}
\titleformat{\subsection}
  {\normalfont\sffamily\large\bfseries}
  {}{0pt}{}
\titleformat{\subsubsection}
  {\normalfont\sffamily\bfseries}
  {}{0pt}{}
\titleformat*{\paragraph}
  {\sffamily\normalsize}

\usepackage{fancyhdr}
\makeatletter
\let\ps@plain\ps@fancy
\fancyheadoffset[L]{4.5cm}
\fancyfootoffset[L]{4.5cm}

\definecolor{linky}{rgb}{0.0, 0.5, 1.0}

\newtcolorbox{repobox}
   {colback=red, colframe=red!75!black,
     boxrule=0.5pt, arc=2pt, left=6pt, right=6pt, top=3pt, bottom=3pt}

\newcommand{\ExternalLink}{%
   \tikz[x=1.2ex, y=1.2ex, baseline=-0.05ex]{%
       \begin{scope}[x=1ex, y=1ex]
           \clip (-0.1,-0.1)
               --++ (-0, 1.2)
               --++ (0.6, 0)
               --++ (0, -0.6)
               --++ (0.6, 0)
               --++ (0, -1);
           \path[draw,
               line width = 0.5,
               rounded corners=0.5]
               (0,0) rectangle (1,1);
       \end{scope}
       \path[draw, line width = 0.5] (0.5, 0.5)
           -- (1, 1);
       \path[draw, line width = 0.5] (0.6, 1)
           -- (1, 1) -- (1, 0.6);
       }
   }

\patchcmd{\@maketitle}{center}{flushleft}{}{}
\patchcmd{\@maketitle}{center}{flushleft}{}{}
\patchcmd{\@maketitle}{\LARGE}{\LARGE\sffamily}{}{}
\def\maketitle{{%
  
  \AB@maketitle}}
\makeatletter
\renewcommand\AB@affilsepx{ \protect\Affilfont}
\renewcommand\AB@affilnote[1]{{\bfseries #1}\hspace{3pt}}
\renewcommand{\affil}[2][]%
   {\newaffiltrue\let\AB@blk@and\AB@pand
      \if\relax#1\relax\def\AB@note{\AB@thenote}\else\def\AB@note{#1}%
        \setcounter{Maxaffil}{0}\fi
        \begingroup
        \let\href=\href@Orig
        \let\texttt=\textttOrig
        \let\protect\@unexpandable@protect
        \def\thanks{\protect\thanks}\def\footnote{\protect\footnote}%
        \@temptokena=\expandafter{\AB@authors}%
        {\def\\{\protect\\\protect\Affilfont}\xdef\AB@temp{#2}}%
         \xdef\AB@authors{\the\@temptokena\AB@las\AB@au@str
         \protect\\[\affilsep]\protect\Affilfont\AB@temp}%
         \gdef\AB@las{}\gdef\AB@au@str{}%
        {\def\\{, \ignorespaces}\xdef\AB@temp{#2}}%
        \@temptokena=\expandafter{\AB@affillist}%
        \xdef\AB@affillist{\the\@temptokena \AB@affilsep
          \AB@affilnote{\AB@note}\protect\Affilfont\AB@temp}%
      \endgroup
       \let\AB@affilsep\AB@affilsepx
}
\makeatother

\renewcommand\Affilfont{\sffamily\small\mdseries}
\ifnum 0\ifxetex 1\fi\ifluatex 1\fi=0 % if pdftex
  \usepackage[T1]{fontenc}
  \usepackage[utf8]{inputenc}

\else % if luatex or xelatex
  \ifxetex
    \usepackage{mathspec}
    \usepackage{fontspec}

  \else
    \usepackage{fontspec}
  \fi
  \defaultfontfeatures{Ligatures=TeX,Scale=MatchLowercase}

\fi
\IfFileExists{upquote.sty}{\usepackage{upquote}}{}
\IfFileExists{microtype.sty}{%
\usepackage{microtype}
\UseMicrotypeSet[protrusion]{basicmath} % disable protrusion for tt fonts
}{}

\usepackage{hyperref}
\hypersetup{unicode=true,
            pdftitle={FAT Forensics: A Python Toolbox for Implementing and Deploying Fairness, Accountability and Transparency Algorithms in Predictive Systems},
            pdfborder={0 0 0},
            breaklinks=true}
\let\addcontentslineOrig=\addcontentsline
\def\addcontentsline#1#2#3{\bgroup
  \let\texttt=\textttOrig\addcontentslineOrig{#1}{#2}{#3}\egroup}
\let\markbothOrig\markboth
\def\markboth#1#2{\bgroup
  \let\texttt=\textttOrig\markbothOrig{#1}{#2}\egroup}
\let\markrightOrig\markright
\def\markright#1{\bgroup
  \let\texttt=\textttOrig\markrightOrig{#1}\egroup}

\usepackage{graphicx,grffile}
\makeatletter
\def\maxwidth{\ifdim\Gin@nat@width>\linewidth\linewidth\else\Gin@nat@width\fi}
\def\maxheight{\ifdim\Gin@nat@height>\textheight\textheight\else\Gin@nat@height\fi}
\makeatother
\setkeys{Gin}{width=\maxwidth,height=\maxheight,keepaspectratio}
\IfFileExists{parskip.sty}{%
\usepackage{parskip}
}{% else
\setlength{\parindent}{0pt}
\setlength{\parskip}{6pt plus 2pt minus 1pt}
}
\providecommand{\tightlist}{%
  \setlength{\itemsep}{0pt}\setlength{\parskip}{0pt}}
\ifx\paragraph\undefined\else
\let\oldparagraph\paragraph
\renewcommand{\paragraph}[1]{\oldparagraph{#1}\mbox{}}
\fi
\ifx\subparagraph\undefined\else
\let\oldsubparagraph\subparagraph
\renewcommand{\subparagraph}[1]{\oldsubparagraph{#1}\mbox{}}
\fi

\title{FAT Forensics: A Python Toolbox for Implementing and Deploying
Fairness, Accountability and Transparency Algorithms in Predictive
Systems}

        \author[1]{Kacper Sokol}
          \author[2]{Alexander Hepburn}
          \author[2]{Rafael Poyiadzi}
          \author[2]{Matthew Clifford}
          \author[2]{Raul Santos-Rodriguez}
          \author[1]{Peter Flach}
    
      \affil[1]{Department of Computer Science, University of Bristol}
      \affil[2]{Department of Engineering Mathematics, University of
Bristol}
  \date{\vspace{-7ex}}

\begin{document}
\maketitle

\marginpar{

  \begin{flushleft}
  %\hrule
  \sffamily\small

  {\bfseries DOI:} \href{https://doi.org/10.21105/joss.01904}{\color{linky}{10.21105/joss.01904}}

  \vspace{2mm}

  {\bfseries Software}
  \begin{itemize}
    \setlength\itemsep{0em}
    \item \href{https://github.com/openjournals/joss-reviews/issues/1904}{\color{linky}{Review}} \ExternalLink
    \item \href{https://github.com/fat-forensics/fat-forensics}{\color{linky}{Repository}} \ExternalLink
    \item \href{https://doi.org/10.5281/zenodo.3833199}{\color{linky}{Archive}} \ExternalLink
  \end{itemize}

  \vspace{2mm}

  \par\noindent\hrulefill\par

  \vspace{2mm}

  {\bfseries Editor:} \href{http://arokem.org/}{Ariel Rokem} \ExternalLink \\
  \vspace{1mm}
    {\bfseries Reviewers:}
  \begin{itemize}
  \setlength\itemsep{0em}
    \item \href{https://github.com/bernease}{@bernease}
    \item \href{https://github.com/osolari}{@osolari}
    \end{itemize}
    \vspace{2mm}

  {\bfseries Submitted:} 12 September 2019\\
  {\bfseries Published:} 19 May 2020

  \vspace{2mm}
  {\bfseries License}\\
  Authors of papers retain copyright and release the work under a Creative Commons Attribution 4.0 International License (\href{http://creativecommons.org/licenses/by/4.0/}{\color{linky}{CC BY 4.0}}).

  \end{flushleft}
}

\hypertarget{background}{%
\section{Background}\label{background}}

Predictive systems, in particular machine learning algorithms, can take
important, and sometimes legally binding, decisions about our everyday
life. In most cases, however, these systems and decisions are neither
regulated nor certified. Given the potential harm that these algorithms
can cause, their qualities such as \textbf{fairness},
\textbf{accountability} and \textbf{transparency} (FAT) are of paramount
importance. To ensure high-quality, fair, transparent and reliable
predictive systems, we developed an open source Python package called
\emph{FAT Forensics}. It can inspect important fairness, accountability
and transparency aspects of predictive algorithms to automatically and
objectively report them back to engineers and users of such systems. Our
toolbox can evaluate all elements of a predictive pipeline: data (and
their features), models and predictions. Published under the BSD
3-Clause open source licence, \emph{FAT Forensics} is opened up for
personal and commercial usage.

\hypertarget{summary}{%
\section{Summary}\label{summary}}

\emph{FAT Forensics} is designed as an interoperable framework for
\emph{implementing}, \emph{testing} and \emph{deploying} novel
algorithms devised by the FAT research community. It facilitates their
evaluation and comparison against the state of the art, thereby
democratising access to these techniques. In addition to supporting
research in this space, the toolbox is capable of analysing all
components of a predictive pipeline -- data, models and predictions --
by considering their fairness, accountability (robustness, security,
safety and privacy) and transparency (interpretability and
explainability).

\emph{FAT Forensics} collates all of these diverse tools and algorithms
under a common application programming interface. This is achieved with
a modular design that allows to share and reuse a collection of core
algorithmic components -- see Figure 1. This architecture makes the
process of creating new algorithms as easy as connecting the right
blocks, therefore supporting a range of diverse use cases.

\begin{figure}
\centering
\includegraphics[width=0.45\textwidth,height=\textheight]{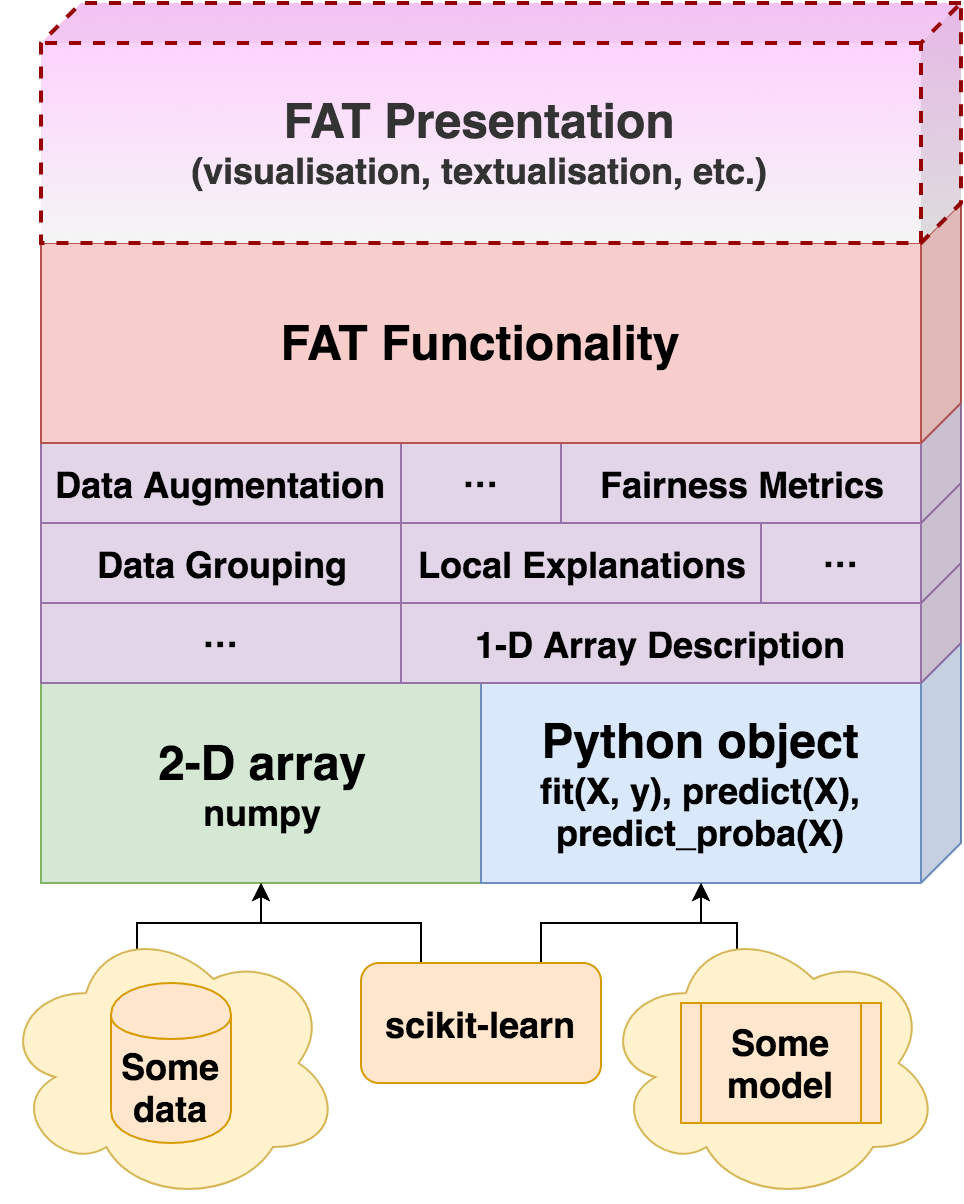}
\caption{Modular architecture of FAT Forensics.}
\end{figure}

The format requirements for data sets and predictive models are kept to
a minimum, lowering any barriers for adoption of \emph{FAT Forensics} in
new and already well-established projects. In this abstraction a data
set is assumed to be a two-dimensional NumPy array: either a classic or
a structured array. The latter is a welcome addition given that some of
the features may be categorical (string-based). A predictive model is
assumed to be a plain Python object that has \texttt{fit},
\texttt{predict} and, optionally, \texttt{predict\_proba} methods. This
flexibility makes our package compatible with scikit-learn (Pedregosa et
al., 2011) -- the most popular Python machine learning toolbox --
without introducing additional dependencies. Moreover, this approach
makes \emph{FAT Forensics} compatible with other packages for predictive
modelling since their predictive functions can be easily wrapped inside
a Python object with all the required methods.

Our package improves over existing solutions as it collates algorithms
across the FAT domains, taking advantage of their shared functional
building blocks. The common interface layer of the toolbox supports
several \emph{modes of operation}. The \textbf{research mode} (data in
-- visualisation out), where the tool can be loaded into an interactive
Python session, e.g., a Jupyter Notebook, supports prototyping and
exploratory analysis. This mode is intended for FAT researchers who may
use it to propose new fairness metrics, compare them with the existing
ones or use them to inspect a new system or a data set. The
\textbf{deployment mode} (data in -- data out) can be used as a part of
a data processing pipeline to provide a (numerical) FAT analytics,
supporting automated reporting and dashboarding. This mode is intended
for machine learning engineers and data scientists who may use it to
monitor or evaluate a predictive system during its development and
deployment.

To encourage long-term maintainability, sustainability and
extensibility, \emph{FAT Forensics} has been developed employing
software engineering best practice such as unit testing, continuous
integration, well-defined package structure and consistent code
formatting. Furthermore, our toolbox is supported by a thorough and
beginner-friendly documentation that is based on four main pillars,
which together build up the user's confidence in using the package:

\begin{itemize}
\tightlist
\item
  narrative-driven \textbf{tutorials} designated for new users, which
  provide a step-by-step guidance through practical use cases of all the
  main aspects of the package;
\item
  \textbf{how-to guides} created for relatively new users of the
  package, which showcase the flexibility of the toolbox and explain how
  to use it to solve user-specific FAT challenges; for example, how to
  build your own local surrogate explainer by pairing a data augmenter
  and an inherently transparent local model;
\item
  \textbf{API documentation} describing functional aspects of the
  algorithms implemented in the package and designated for a technical
  audience as a reference material; it is complemented by task-focused
  \emph{code examples} that put the functions, objects and methods in
  context; and
\item
  a \textbf{user guide} discussing theoretical aspects of the algorithms
  implemented in the package such as their restrictions, caveats,
  computational time and memory complexity, among others.
\end{itemize}

We hope that this effort will encourage the FAT community to contribute
their algorithms to \emph{FAT Forensics}. We offer it as an attractive
alternative to releasing yet more standalone packages, keeping the
toolbox at the frontiers of algorithmic fairness, accountability and
transparency research. For a more detailed description of \emph{FAT
Forensics}, we point the reader to its documentation\footnote{https://fat-forensics.org}
and the paper (Sokol et al., 2019) describing its design, scope and
usage examples.

\hypertarget{related-work}{%
\section{Related Work}\label{related-work}}

A recent attempt to create a common framework for FAT algorithms is the
\emph{What-If} tool\footnote{https://pair-code.github.io/what-if-tool},
which implements various fairness and explainability approaches. A
number of Python packages collating multiple state-of-the-art algorithms
for either fairness, accountability or transparency also exist.
Available algorithmic \emph{transparency} packages include:

\begin{itemize}
\tightlist
\item
  Skater\footnote{https://github.com/oracle/Skater} (Kramer et al.,
  2018),
\item
  ELI5\footnote{https://github.com/TeamHG-Memex/eli5},
\item
  Microsoft's Interpret\footnote{https://github.com/interpretml/interpret}
  (Nori et al., 2019), and
\item
  IBM's AI Explainability 360\footnote{https://github.com/IBM/AIX360}.
\end{itemize}

Packages implementing individual algorithms are also popular. For
example, LIME\footnote{https://github.com/marcotcr/lime} for Local
Interpretable Model-agnostic Explanations (Ribeiro et al., 2016) and
PyCEbox\footnote{https://github.com/AustinRochford/PyCEbox} for Partial
Dependence (Friedman, 2001) and Individual Conditional Expectation
(Goldstein et al., 2015) plots.

Algorithmic \emph{fairness} packages are also ubiquitous, for example:
Microsoft's fairlearn\footnote{https://github.com/fairlearn/fairlearn}
(Agarwal et al., 2018) and IBM's AI Fairness 360\footnote{https://github.com/IBM/AIF360}
(Bellamy et al., 2018). However, \emph{accountability} is relatively
underexplored. The most prominent software in this space deals with
robustness of predictive systems against adversarial attacks, for
example:

\begin{itemize}
\tightlist
\item
  FoolBox\footnote{https://github.com/bethgelab/foolbox},
\item
  CleverHans\footnote{https://github.com/tensorflow/cleverhans} and
\item
  IBM's Adversarial Robustness 360 Toolbox\footnote{https://github.com/IBM/adversarial-robustness-toolbox}.
\end{itemize}

\emph{FAT Forensics} aims to bring together all of this functionality
from across fairness, accountability and transparency domains with its
modular implementation. This design principle enables the toolbox to
support two modes of operation: research and deployment. Therefore, the
package caters to a diverse audience and supports a range of tasks such
as implementing, testing and deploying FAT solutions. Abstracting away
from fixed data set and predictive model formats adds to its
versatility. The development of the toolbox adheres to best practices
for software engineering and the package is supported by a rich
documentation, both of which make it stand out amongst its peers.

\hypertarget{acknowledgements}{%
\section{Acknowledgements}\label{acknowledgements}}

This work was financially supported by Thales, and is the result of a
collaborative research agreement between Thales and the University of
Bristol.

\hypertarget{references}{%
\section*{References}\label{references}}
\addcontentsline{toc}{section}{References}

\hypertarget{refs}{}
\begin{CSLReferences}{1}{0}
\leavevmode\hypertarget{ref-agarwal2018reductions}{}%
Agarwal, A., Beygelzimer, A., Dudik, M., Langford, J., \& Wallach, H.
(2018). A reductions approach to fair classification. In J. Dy \& A.
Krause (Eds.), \emph{Proceedings of the 35th international conference on
machine learning} (Vol. 80, pp. 60--69). PMLR.
\url{http://proceedings.mlr.press/v80/agarwal18a.html}

\leavevmode\hypertarget{ref-bellamy2018ai}{}%
Bellamy, R. K. E., Dey, K., Hind, M., Hoffman, S. C., Houde, S., Kannan,
K., Lohia, P., Martino, J., Mehta, S., Mojsilovic, A., Nagar, S.,
Ramamurthy, K. N., Richards, J., Saha, D., Sattigeri, P., Singh, M.,
Varshney, K. R., \& Zhang, Y. (2018). AI {F}airness 360: An extensible
toolkit for detecting, understanding, and mitigating unwanted
algorithmic bias. \emph{arXiv Preprint arXiv:1810.01943}.

\leavevmode\hypertarget{ref-friedman2001greedy}{}%
Friedman, J. H. (2001). Greedy function approximation: A gradient
boosting machine. \emph{Annals of Statistics}, 1189--1232.
\url{https://doi.org/10.1214/aos/1013203451}

\leavevmode\hypertarget{ref-goldstein2015peeking}{}%
Goldstein, A., Kapelner, A., Bleich, J., \& Pitkin, E. (2015). Peeking
inside the black box: Visualizing statistical learning with plots of
individual conditional expectation. \emph{Journal of Computational and
Graphical Statistics}, \emph{24}(1), 44--65.
\url{https://doi.org/10.1080/10618600.2014.907095}

\leavevmode\hypertarget{ref-skater}{}%
Kramer, A., Choudhary, P., silversurfer84, Dyke, B. V., Thai, A.,
Pasumarthy, N., Lemaitre, G., Thompson, D., \& Cook, B. (2018).
\emph{Datascienceinc/skater: 1.1.2} (Version v1.1.2) {[}Computer
software{]}. Zenodo. \url{https://doi.org/10.5281/zenodo.1423046}

\leavevmode\hypertarget{ref-nori2019interpretml}{}%
Nori, H., Jenkins, S., Koch, P., \& Caruana, R. (2019). InterpretML: A
unified framework for machine learning interpretability. \emph{arXiv
Preprint arXiv:1909.09223}.

\leavevmode\hypertarget{ref-scikit-learn}{}%
Pedregosa, F., Varoquaux, G., Gramfort, A., Michel, V., Thirion, B.,
Grisel, O., Blondel, M., Prettenhofer, P., Weiss, R., Dubourg, V.,
Vanderplas, J., Passos, A., Cournapeau, D., Brucher, M., Perrot, M., \&
Duchesnay, E. (2011). Scikit-learn: Machine learning in {P}ython.
\emph{Journal of Machine Learning Research}, \emph{12}, 2825--2830.

\leavevmode\hypertarget{ref-ribeiro2016why}{}%
Ribeiro, M. T., Singh, S., \& Guestrin, C. (2016). "Why should {I} trust
you?": Explaining the predictions of any classifier. \emph{Proceedings
of the 22nd {ACM} {SIGKDD} International Conference on Knowledge
Discovery and Data Mining, San Francisco, CA, USA, August 13-17, 2016},
1135--1144. \url{https://doi.org/10.18653/v1/n16-3020}

\leavevmode\hypertarget{ref-sokol2022fatf}{}%
Sokol, K., Santos-Rodriguez, R., \& Flach, P. (2022). {FAT Forensics}:
{A} {P}ython toolbox for algorithmic fairness, accountability and
transparency. \emph{Software Impacts}, \emph{14}, 100406.
\url{https://doi.org/10.1016/j.simpa.2022.100406}

\end{CSLReferences}

\end{document}